\documentclass[11pt,a4paper]{article}
\usepackage[hyperref]{acl2018}
\usepackage{times}
\usepackage{latexsym}

\usepackage{url}
\usepackage{caption}
\usepackage{hyperref}       
\usepackage{url}            
\usepackage{booktabs}       
\usepackage{amsfonts}       
\usepackage{nicefrac}       
\usepackage{microtype}      

\usepackage{tabularx}
\usepackage{amsmath,amssymb,amsthm}
\usepackage{graphicx}
\usepackage{subcaption}
\usepackage{listings}
\usepackage{upquote}
\usepackage{wrapfig}
\usepackage{lipsum}
\usepackage{array}
\usepackage{multicol}
\usepackage{multirow}
\usepackage{url}
\usepackage{color}
\usepackage{wrapfig,lipsum,booktabs}
\usepackage{makecell}
\usepackage{booktabs}

\usepackage{tabulary}
\newcolumntype{K}[1]{>{\centering\arraybackslash}p{#1}}

\aclfinalcopy 

\title{Speeding up Context-based Sentence Representation Learning \\with Non-autoregressive Convolutional Decoding}
\author{Shuai Tang$^1$,  Hailin Jin$^2$,  Chen Fang$^2$,  Zhaowen Wang$^2$, Virginia R. de Sa$^1$  \\
$^1$ Department of Cognitive Science, UC San Diego \\
$^2$ Adobe Research \\
\texttt{\{shuaitang93,desa\}@ucsd.edu, \{hljin,cfang,zhawang\}@adobe.com} \\
}

\date{}

\begin{document}
\maketitle
\begin{abstract}
Context plays an important role in human language understanding, thus it may also be useful for machines learning vector representations of language. In this paper, we explore an asymmetric encoder-decoder structure for unsupervised context-based sentence representation learning. We carefully designed experiments to show that neither an autoregressive decoder nor an RNN decoder is required. After that, we designed a model which still keeps an RNN as the encoder, while using a non-autoregressive convolutional decoder. We further combine a suite of effective designs to significantly improve model efficiency while also achieving better performance. Our model is trained on two different large unlabelled corpora, and in both cases the transferability is evaluated on a set of downstream NLP tasks. We empirically show that our model is simple and fast while producing rich sentence representations that excel in downstream tasks.

\end{abstract}

\section{Introduction}

Learning distributed representations of sentences is an important and hard topic in both the deep learning and natural language processing communities, since it requires machines to encode a sentence with rich language content into a fixed-dimension vector filled with real numbers. Our goal is to build a distributed sentence encoder learnt in an unsupervised fashion by exploiting the structure and relationships in a large unlabelled corpus. 

Numerous studies in human language processing have supported that rich semantics of a word or  sentence can be inferred from its context \citep{harris1954distributional,firth57synopsis}. The idea of learning from the co-occurrence \citep{Turney2010FromFT} was recently successfully applied to vector representation learning for words in \citet{Mikolov2013DistributedRO} and \citet{Pennington2014GloveGV}. 

A very recent successful application of the distributional hypothesis \citep{harris1954distributional} at the sentence-level is the skip-thoughts model \citep{Kiros2015SkipThoughtV}. The skip-thoughts model learns to encode the current sentence and decode the surrounding two sentences instead of the input sentence itself, which achieves overall good performance on all tested downstream NLP tasks that cover various topics. The major issue is that the training takes too long since there are two RNN decoders to reconstruct the previous sentence and the next one independently. Intuitively, given the current sentence, inferring the previous sentence and inferring the next one should be different, which supports the usage of two independent decoders in the skip-thoughts model.
However, \citet{Tang2017Rethinking} proposed the skip-thought neighbour model, which only decodes the next sentence based on the current one, and has similar performance on downstream tasks compared to that of their implementation of the skip-thoughts model.



In the encoder-decoder models for learning sentence representations, only the encoder will be used to map sentences to vectors after training, which implies that the quality of the generated language is not our main concern. This leads to our two-step experiment to check the necessity of applying an autoregressive model as the decoder. In other words, since the decoder's performance on language modelling is not our main concern, it is  preferred to reduce the complexity of the decoder to speed up the training process. In our experiments, the first step is to check whether ``teacher-forcing'' is required during training if we stick to using an autoregressive model as the decoder, and the second step is to check whether an autoregressive decoder is necessary to learn a good sentence encoder. Briefly, the experimental results show that an autoregressive decoder is indeed not essential in learning a good sentence encoder; thus the two findings of our experiments lead to our final model design.

Our proposed model has an asymmetric encoder-decoder structure, which keeps an RNN as the encoder and has a CNN as the decoder, and the model explores using only the subsequent context information as the supervision. The asymmetry in both model architecture and training pair reduces a large amount of the training time.

The contribution of our work is summarised as:
\begin{enumerate}
\item We design experiments to show that an autoregressive decoder or an RNN decoder is not necessary in the encoder-decoder type of models for learning sentence representations, and based on our results, we present two findings. \textbf{Finding I}: It is not necessary to input the correct words into an autoregressive decoder for learning sentence representations. \textbf{Finding II}: The model with an autoregressive decoder performs similarly to the model with a predict-all-words decoder.
\item The two findings above lead to our final model design, which keeps an RNN encoder while using a CNN decoder and learns to encode the current sentence and decode the subsequent contiguous words all at once.
\item With a suite of techniques, our model performs decently on the downstream tasks, and can be trained efficiently on a large unlabelled corpus.
\end{enumerate}

The following sections will introduce the components in our ``RNN-CNN'' model, and discuss our experimental design. 


\section{RNN-CNN Model}
\label{rnn-cnn}
\begin{figure*}[t]
\begin{center}
\includegraphics[width=0.8\textwidth]{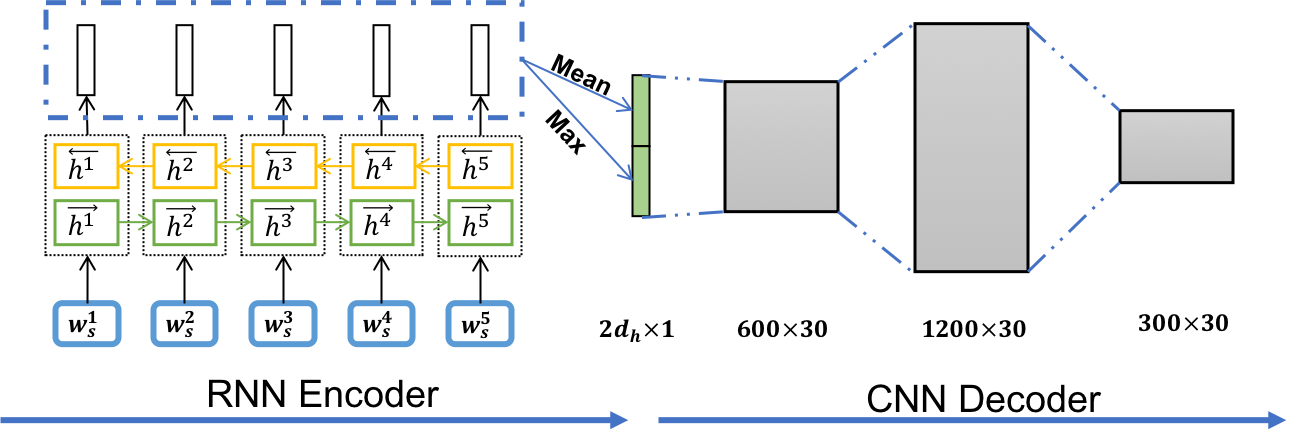}
\end{center}
\caption{Our proposed model is composed of an RNN encoder, and a CNN decoder. During training, a batch of sentences are sent to the model, and the RNN encoder computes a vector representation for each of sentences; then the CNN decoder needs to reconstruct the paired target sequence, which contains 30 contiguous words right after the input sentence, given the vector representation. $300$ is the dimension of word vectors. $2d_h$ is the dimension of the sentence representation which varies with the RNN encoder size. (Better view in colour.)}
\label{model}
\end{figure*}

Our model is highly asymmetric in terms of both the training pairs and the model structure. Specifically, our model has an RNN as the encoder, and a CNN as the decoder. During training, the encoder takes the $i$-th sentence $S_i$ as the input, and then produces a fixed-dimension vector $\mathbf{z}_i$ as the sentence representation; the decoder is applied to reconstruct the paired target sequence $T_i$ that contains the subsequent contiguous words. The distance between the generated sequence and the target one is measured by the cross-entropy loss at each position in $T_i$. An illustration is in Figure \ref{model}. (For simplicity, we omit the subscript $i$ in this section.) 

\textbf{1. Encoder:} The encoder is a bi-directional Gated Recurrent Unit (GRU, \citet{Chung2014EmpiricalEO})\footnote{We experimented with both Long-short Term Memory (LSTM, \citet{Hochreiter1997LongSM}) and GRU. Since LSTM didn't give us significant performance boost, and generally GRU runs faster than LSTM, in our experiments, we stick to using GRU in the encoder.}. Suppose that an input sentence $S$ contains $M$ words, which are $\{w^1_s,w^2_s,...,w^M_s\}$, and they are transformed by an embedding matrix $\mathbf{E}\in\mathbb{R}^{d_e\times |V|}$ to word vectors\footnote{$V$ is the vocabulary, and $|V|$ is the number of unique words in the vocabulary. $d_e$ is the dimension of each word vector.}. The bi-directional GRU takes one word vector at a time, and processes the input sentence in both the forward and backward directions; both sets of hidden states are concatenated to form the hidden state matrix $\mathbf{H} = [\mathbf{h}^1,\mathbf{h}^2,...,\mathbf{h}^M] \in \mathbb{R}^{d_h\times M}$, where $d_h$ is the dimension of the hidden states $\mathbf{h}^{m}=\left[\overleftarrow{\mathbf{h}^m};\overrightarrow{\mathbf{h}^m}\right]$ ($\forall m \in \{1,2,...,M\}$).  

\textbf{2. Representation: } We aim to provide a model with faster training speed and better transferability than existing algorithms; thus we choose to apply a parameter-free composition function, which is a concatenation of the outputs from a global mean pooling over time and a global max pooling over time, on the computed sequence of hidden states $\mathbf{H}$. The composition function is represented as
\begin{align}
\mathbf{z} = \left[\frac{1}{M}\sum_{m=1}^M\mathbf{h}^m;\textrm{MaxPool}(\mathbf{H})\right]
\end{align}

where $\textrm{MaxPool}$ is the max operation on  each row of the matrix $\mathbf{H}$, which outputs a vector with dimension $d_h$. Thus the representation $\mathbf{z} \in \mathbb{R}^{2d_h}$. 

\textbf{3. Decoder: } The decoder is a 3-layer CNN to reconstruct the paired target sequence $T$, which needs to expand $\mathbf{z}$, which can be considered as a sequence with only one element, to a sequence with $T$ elements. Intuitively, the decoder could be a stack of deconvolution layers. For fast training speed, we optimised the architecture to make it possible to use fully-connected layers and convolution layers in the decoder, since generally, convolution layers run faster than deconvolution layers in modern deep learning frameworks. 

Suppose that the target sequence $T$ has $N$ words, which are $\{w^1_t,w^2_t,...,w^N_t\}$, the first layer of deconvolution will expand $\mathbf{z}\in\mathbb{R}^{2d_h\times 1}$, into a feature map with $N$ elements. It can be easily implemented as a concatenation of outputs from $N$ linear transformations in parallel. Then the second and third layer are 1D-convolution layers. The output feature map is $\mathbf{U}=[\mathbf{u}^1,\mathbf{u}^2,...,\mathbf{u}^{N}]\in\mathbb{R}^{d_e\times N}$, where $d_e$ is the dimension of the word vectors.

Note that our decoder is not an autoregressive model and  has high training efficiency. We will discuss the reason for choosing this decoder which we call a predict-all-words CNN decoder. 

\textbf{4. Objective: } The training objective is to maximise the likelihood of the target sequence being generated from the decoder. Since in our model, each word is predicted independently, a softmax layer is applied after the decoder to produce a probability distribution over words in $V$ at each position, thus the probability of generating a word $w_t^n$ in the target sequence is defined as:
\begin{align}
P(w_t^n)=\frac{e^{\mathbf{E}(w_t^n)^\top\mathbf{u}^n}}{\sum_{w\in V}e^{\mathbf{E}(w)^\top\mathbf{u}^n}}, 
\end{align}
where, $\mathbf{E}(w)$ is the vector representation of $w$ in the embedding matrix $\mathbf{E}$, and $\mathbf{E}(w)^\top\mathbf{u}^n$ is the dot-product between the word vector and the feature vector produced by the decoder at position $n$ . The training objective is to minimise the sum of the negative log-likelihood over all positions in the target sequence $T$:
\begin{align}
&\mathcal{L}(\boldsymbol{\phi},\boldsymbol{\theta}) = -\log P(T|S;\boldsymbol{\phi},\boldsymbol{\theta}) \nonumber \\ 
&= -\sum_{n=1}^N\log P(w^n_t|w^1_s,w^2_s,...,w^M_s;\boldsymbol{\phi},\boldsymbol{\theta}), 
\end{align}
where $\boldsymbol{\phi}$ and $\boldsymbol{\theta}$ contain the parameters in the encoder and the decoder, respectively. The training objective $\mathcal{L}(\boldsymbol{\phi},\boldsymbol{\theta})$ is summed over all sentences in the training corpus.

\begin{table*}[t]
\fontsize{8}{11}\selectfont
\begin{center}
\begin{tabular}{ c | c c c | c | c c }
\specialrule{2pt}{1pt}{1pt}
\textbf{Decoder} & \textbf{SICK-R} & \textbf{SICK-E} & \textbf{STS14} & \textbf{MSRP} (Acc/F1) &    \textbf{SST} & \textbf{TREC} \\
\specialrule{2pt}{1pt}{1pt}
\multicolumn{7}{c}{\textbf{auto-regressive RNN as decoder}}\\
\hline
Teacher-Forcing & 0.8530 & 82.6 & 0.51/0.50 & 74.1 / 81.7 & 82.5 & 88.2\\
Always Sampling & 0.8576 & 83.2 & 0.55/0.53 & 74.7 / 81.3 & 80.6 & 87.0\\
Uniform Sampling & 0.8559 & 82.9 & 0.54/0.53 & 74.0 / 81.8 & 81.0 & 87.4\\
\specialrule{2pt}{1pt}{1pt}
\multicolumn{7}{c}{\textbf{auto-regressive CNN as decoder}}\\
\hline
Teacher-Forcing & 0.8510 & 82.8 & 0.49/0.48 & 74.7 / 82.8 & 81.4 & 82.6 \\
Always Sampling & 0.8535 & 83.3 & 0.53/0.52 & 75.0 / 81.7 & 81.4 & 87.6 \\
Uniform Sampling & 0.8568 & 83.4 & 0.56/0.54 & 74.7 / 81.4 & 83.0 & 88.4 \\
\specialrule{2pt}{1pt}{1pt}
\multicolumn{7}{c}{\textbf{predict-all-words RNN as decoder}}\\
\hline
RNN & 0.8508 & 82.8 & 0.58/0.55 & 74.2 / 82.8 & 81.6 & 88.8 \\
\specialrule{2pt}{1pt}{1pt}
\multicolumn{7}{c}{\textbf{predict-all-words CNN as decoder}}\\
\hline
CNN & 0.8530 & 82.6 & \textbf{0.58}/\textbf{0.56} & \textbf{75.6} / 82.9 & 82.8 & 89.2 \\
CNN-MaxOnly & 0.8465 & 82.6 & 0.50/0.47 & 73.3 / 81.5 & 79.1 & 82.2 \\ 
\hline
\multicolumn{7}{c}{Double-sized RNN Encoder} \\
\hline
CNN & \textbf{0.8631} & \textbf{83.9} & \textbf{0.58}/0.55 & 74.7 / \textbf{83.1} & \textbf{83.4} & \textbf{90.2} \\
CNN-MaxOnly & 0.8485 & 83.2 &0.47/0.44 & 72.9 / 80.8 & 82.2 & 86.6 \\
\specialrule{2pt}{1pt}{1pt}
\end{tabular}
\end{center}
\caption{The models here all have a bi-directional GRU as the encoder (dimensionality 300 in each direction). The default way of producing the representation is a concatenation of outputs from a global \emph{mean-pooling} and a global \emph{max-pooling}, while ``$\cdot$\emph{-MaxOnly}'' refers to the model with only global max-pooling. Bold numbers are the best results among all presented models. We found that \textbf{1)} inputting correct words to an autoregressive decoder is not necessary; \textbf{2)} predict-all-words decoders work roughly the same as autoregressive decoders; \textbf{3)} mean+max pooling provides stronger transferability than the max-pooling alone does. The table supports our choice of the predict-all-words CNN decoder and the way of producing vector representations from the bi-directional RNN encoder.}
\label{autoregressive}
\end{table*}

\section{Architecture Design}

We use an encoder-decoder model and use context for learning sentence representations in an unsupervised fashion. Since the decoder won't be used after training, and the quality of the generated sequences is not our main focus, it is important to study the design of the decoder. Generally, a fast training algorithm is preferred; thus proposing a new decoder with high training efficiency and also strong transferability is crucial for an encoder-decoder model.

\subsection{CNN as the decoder}

Our design of the decoder is basically a 3-layer ConvNet that predicts all words in the target sequence at once. In contrast, existing work, such as skip-thoughts \cite{Kiros2015SkipThoughtV}, and CNN-LSTM \cite{Gan2017LearningGS}, use autoregressive RNNs as the decoders. As known, an autoregressive model is good at generating sequences with high quality, such as language and speech. However, an autoregressive decoder seems to be unnecessary in an encoder-decoder model for learning sentence representations, since it won't be used after training, and it takes up a large portion of the training time to compute the output and the gradient. Therefore, we conducted experiments to test the necessity of using an autoregressive decoder in learning sentence representations, and we had two findings.

\textbf{Finding I: It is not necessary to input the correct words into an autoregressive decoder for learning sentence representations.} 

The experimental design was inspired by \citet{Bengio2015ScheduledSF}. The model we designed for the experiment has a bi-directional GRU as the encoder, and an autoregressive decoder, including both RNN and CNN. We started by analysing the effect of different sampling strategies of the input words on learning an auto-regressive decoder.

We compared three sampling strategies of input words in decoding the target sequence with an autoregressive decoder: (1) \textbf{Teacher-Forcing}: the decoder always gets the ground-truth words; (2) \textbf{Always Sampling}: at time step $t$, a word is sampled from the multinomial distribution predicted at  time step $t-1$; (3) \textbf{Uniform Sampling}: a word is uniformly sampled from the dictionary $V$, then fed to the decoder at every time step.

The results are presented in Table \ref{autoregressive} (top two subparts). As we can see, the three decoding settings do not differ significantly in terms of the performance on selected downstream tasks, with RNN or CNN as the decoder. The results show that, in terms of learning good sentence representations, the autoregressive decoder doesn't require the correct ground-truth words as the inputs. 

\textbf{Finding II: The model with an autoregressive decoder performs similarly to the model with a predict-all-words decoder.}

With \textbf{Finding I}, we conducted an experiment to test whether the model needs an autoregressive decoder at all. In this experiment, the goal is to compare the performance of the predict-all-words decoders and that of the autoregressive decoders separate from the RNN/CNN distinction, thus we designed a predict-all-words CNN decoder and RNN decoder. The predict-all-words CNN decoder is described in Section \ref{rnn-cnn}, which is a stack of three convolutional layers, and all words are predicted once at the output layer of the decoder. The predict-all-words RNN decoder is built based on our CNN decoder. To keep the number of parameters of the two predict-all-words decoder roughly the same, we replaced the last two convolutional layers with a bidirectional GRU.

The results are also presented in Table \ref{autoregressive} (3rd and 4th subparts). The performance of the predict-all-words RNN decoder does not significantly differ from that of any one of the autoregressive RNN decoders, and the same situation can be also observed in CNN decoders. 

These two findings indeed support our choice of using a predict-all-words CNN as the decoder, as it brings the model high training efficiency while maintaining strong transferability.

\subsection{Mean+Max Pooling}

Since the encoder is a bi-directional RNN in our model, we have multiple ways to select/compute on the generated hidden states to produce a sentence representation. Instead of using the last hidden state as the sentence representation as done in skip-thoughts \citep{Kiros2015SkipThoughtV} and SDAE \citep{Hill2016LearningDR}, we followed the idea proposed in \citet{Chen2016EnhancingAC}. They built a model for supervised training on the SNLI dataset \citep{Bowman2015ALA} that concatenates the outputs from a global mean pooling over time and a global max pooling over time to serve as the sentence representation, and showed a performance boost on the SNLI task. \citet{Conneau2017SupervisedLO} found that the model with global max pooling function provides stronger transferability than the model with a global mean pooling function does. 

In our proposed RNN-CNN model, we empirically show that the mean+max pooling provides stronger transferability than the max pooling alone does, and the results are presented in the last two sections of Table \ref{autoregressive}. The concatenation of a mean-pooling and a max pooling function is actually a parameter-free composition function, and the computation load is negligible compared to all the heavy matrix multiplications in the model. Also, the non-linearity of the max pooling function augments the mean pooling function for constructing a representation that captures a more complex composition of the syntactic information. 

\subsection{Tying Word Embeddings and Word Prediction Layer}

We choose to share the parameters in the word embedding layer of the RNN encoder and the word prediction layer of the CNN decoder. Tying was shown in both \citet{Inan2016TyingWV} and \citet{Press2017UsingTO}, and it generally helped to learn a better language model. In our model, tying also drastically reduces the number of parameters, which could potentially prevent overfitting.

Furthermore, we initialise the word embeddings with pretrained word vectors, such as word2vec \citep{Mikolov2013DistributedRO} and GloVe \citep{Pennington2014GloveGV}, since it has been shown that these pretrained word vectors can serve as a good initialisation for deep learning models, and more likely lead to better results than a random initialisation.

\subsection{Study of the Hyperparameters in Our Model Design}

We studied hyperparameters in our model design based on three out of 10 downstream tasks, which are SICK-R, SICK-E  \citep{Marelli2014ASC}, and STS14 \citep{Agirre2014SemEval2014T1}. The first model we created, which is reported in Section \ref{rnn-cnn}, is a decent design, and the following variations didn't give us much performance change except improvements brought by increasing the dimensionality of the encoder. However, we think it is worth mentioning the effect of hyperparameters in our model design. We present the Table \ref{architecture} in the supplementary material and we summarise it as follows:\\
\textbf{1.} Decoding the next sentence performed similarly to decoding the subsequent contiguous words.\\
\textbf{2.} Decoding the subsequent 30 words, which was adopted from the skip-thought training code\footnote{https://github.com/ryankiros/skip-thoughts/}, gave reasonably good performance. More words for decoding didn't give us a significant performance gain, and  took longer to train.\\
\textbf{3.} Adding more layers into the decoder and enlarging the dimension of the convolutional layers indeed sightly improved the performance on the three downstream tasks, but as training efficiency is one of our main concerns, it wasn't worth sacrificing training efficiency for the minor performance gain.\\
\textbf{4.} Increasing the dimensionality of the RNN encoder improved the model performance, and the additional training time required was less than needed for 
increasing the complexity in the CNN decoder. We report results from both smallest and largest models in Table \ref{related-work}.
\begin{table*}[!t]
\fontsize{9}{12}\selectfont
\begin{center}
\begin{tabular} {l| K{0.4cm} | K{1.2cm} K{1.1cm} K{1.0cm} K{1.1cm} | K{0.6cm} K{0.5cm} K{0.5cm} K{0.5cm} K{0.6cm} K{0.6cm}}
\Xhline{2pt}
\textbf{Model} & \textbf{Hrs} & \textbf{SICK-R} & \textbf{SICK-E} & \textbf{STS14} & \textbf{MSRP} & \textbf{TREC} & \textbf{MR} & \textbf{CR} & \textbf{SUBJ} & \textbf{MPQA} & \textbf{SST} \\
\Xhline{2pt}
Measurement  & & $r$ & Acc. & $r$/$\rho$ & Acc./F1 & \multicolumn{6}{c}{Accuracy}\\
\hline
\multicolumn{12}{r}{\emph{Unsupervised training with unordered sentences}} \\
\hline
ParagraphVec & 4 & - & - & 0.42/0.43 & 72.9/81.1 & 59.4 & 60.2 & 66.9 & 76.3 & 70.7 & - \\
word2vec BOW & 2 & 0.8030 & 78.7 & 0.65/\textbf{0.64} & 72.5/81.4 & 83.6 & 77.7 & \textbf{79.8} & 90.9 & \textbf{88.3} &  79.7 \\
fastText BOW & - & 0.8000 & 77.9 & 0.63/0.62 & 72.4/81.2 & 81.8 & 76.5 & 78.9 & \textbf{91.6} & 87.4 & 78.8 \\
SIF (GloVe+WR) & - & \textbf{0.8603} & \textbf{84.6} & \textbf{0.69}/  -  & - / - & - & - & - & - & - & \textbf{82.2} \\
GloVe BOW & - & 0.8000 & 78.6 & 0.54/0.56 & 72.1/80.9 & 83.6 & \textbf{78.7} & 78.5 & \textbf{91.6} & 87.6 & 79.8 \\
SDAE & 72 & - & - & 0.37/0.38 & \textbf{73.7}/80.7 & 78.4 & 74.6 & 78.0 & 90.8 & 86.9 & -  \\
\Xhline{2pt}

\hline
\multicolumn{12}{r}{\emph{Unsupervised training with ordered sentences - BookCorpus}} \\
\hline
FastSent & 2 & - & - & \textbf{0.63}/\textbf{0.64} & 72.2/80.3 & 76.8 & 70.8 & 78.4 & 88.7 & 80.6 & - \\
FastSent+AE & 2 & - & - & 0.62/0.62 & 71.2/79.1 & 80.4 & 71.8 & 76.5 & 88.8 & 81.5 & - \\
\hline
Skip-thoughts & 336 & 0.8580 & 82.3 & 0.29/0.35 & 73.0/82.0& 92.2 & 76.5 & 80.1 & 93.6 & 87.1 & 82.0 \\
Skip-thought+LN & 720 & 0.8580 & 79.5 & 0.44/0.45 & - & 88.4 & 79.4 & \textbf{83.1} & 93.7 & 89.3 & 82.9 \\
combine CNN-LSTM & - & 0.8618 & - & - & \underline{\textbf{76.5}}/\underline{\textbf{83.8}} & \underline{\textbf{92.6}} & 77.8 & 82.1 & 93.6 & \textbf{89.4} & - \\
\hline
\textbf{\emph{small RNN-CNN}}\dag & 20 & 0.8530 & 82.6  & 0.58/0.56 & 75.6/82.9 & 89.2 & 77.6 & 80.3 & 92.3 & 87.8 & 82.8 \\
\textbf{\emph{large RNN-CNN}}\dag & 34 & \textbf{0.8698} & \textbf{85.2} & 0.59/0.57 & 75.1/83.2 & 92.2 & \textbf{79.7} & 81.9 & \textbf{94.0} & 88.7 & \textbf{84.1} \\
\Xhline{2pt}
\hline
\multicolumn{12}{r}{\emph{Unsupervised training with ordered sentences - Amazon Book Review}} \\
\hline
\textbf{\emph{small RNN-CNN}}\dag & 21 & 0.8476 & 82.7 & \textbf{0.53}/\textbf{0.53} & 73.8/81.5 & 84.8 & 83.3 & 83.0 & 94.7 & 88.2 & 87.8 \\
\textbf{\emph{large RNN-CNN}}\dag & 33 & \textbf{0.8616} & \textbf{84.3} & 0.51/0.51 & \textbf{75.7}/\textbf{82.8} & \textbf{90.8} & 85.3 & 86.8 & \underline{\textbf{95.3}} & \textbf{89.0} & \underline{\textbf{88.3}} \\
\hline
\hline
\multicolumn{12}{r}{\emph{Unsupervised training with ordered sentences - Amazon Review}} \\
\hline
BYTE m-LSTM  & 720 & 0.7920 & - & - & 75.0/\textbf{82.8} & - & \underline{\textbf{86.9}} & \underline{\textbf{91.4}} & 94.6 & 88.5 & - \\
\Xhline{2pt}
\hline
\multicolumn{12}{r}{\emph{Supervised training - Transfer learning}} \\
\hline
DiscSent & 8 & - & - & - & 75.0/ -  & 87.2 & - & - & 93.0 & - & - \\
DisSent Books 8 & - & 0.8170 & 81.5 & - & -/ -  & 87.2 & \textbf{82.9} & 81.4 & \textbf{93.2} & 90.0 & 80.2 \\
\hline
CaptionRep BOW & 24 & - & - & 0.46/0.42 & -  & 72.2 & 61.9 & 69.3 & 77.4 & 70.8 & - \\
DictRep BOW & 24 & - & - & 0.67/\underline{\textbf{0.70}} & 68.4/76.8 & 81.0 & 76.7 & 78.7 & 90.7 & 87.2 & - \\
InferSent(SNLI) & $<$24 & \underline{\textbf{0.8850}} & 84.6 & 0.68/0.65 & 75.1/82.3 & \textbf{88.7} & 79.9 & 84.6 & 92.1 & 89.8 & 83.3\\
InferSent(AllNLI)  & $<$24 & 0.8840 & \underline{\textbf{86.3}} & \underline{\textbf{0.70}}/0.67 & \textbf{76.2}/\textbf{83.1} & 88.2 & 81.1 & \textbf{86.3} & 92.4 & \underline{\textbf{90.2}} & \textbf{84.6} \\
\Xhline{2pt}
\hline
\end{tabular}
\end{center}
\caption{\textbf{Related Work and Comparison.} As presented, our designed asymmetric RNN-CNN model has strong transferability, and is overall better than existing unsupervised models in terms of fast training speed and good performance on evaluation tasks. ``\dag''s refer to our models, and ``\textbf{small}/\textbf{large}'' refers to the dimension of representation as 1200/4800. \textbf{Bold} numbers are the best ones among the models with same training and transferring setting, and \underline{underlined} numbers are best results among all transfer learning models. The training time of each model was collected from the paper that proposed it.}
\label{related-work}
\end{table*}
\section{Experiment Settings}
\label{settings}

The vocabulary for unsupervised training contains the 20k most frequent words in BookCorpus. In order to generalise the model trained with a relatively small, fixed vocabulary to the much larger set of all possible English words, we followed the vocabulary expansion method proposed in \citet{Kiros2015SkipThoughtV}, which learns a linear mapping from the pretrained word vectors to the learnt RNN word vectors.
Thus, the model benefits from the generalisation ability of the pretrained word embeddings. 

The downstream tasks for evaluation include semantic relatedness (SICK, \citet{Marelli2014ASC}), paraphrase detection (MSRP, \citet{Dolan2004UnsupervisedCO}), question-type classification (TREC, \citet{Li2002LearningQC}), and five benchmark sentiment and subjective datasets, which include movie review sentiment (MR, \citet{Pang2005SeeingSE}, SST, \citet{Socher2013RecursiveDM}), customer product reviews (CR, \citet{Hu2004MiningAS}), subjectivity/objectivity classification (SUBJ, \citet{Pang2004ASE}), opinion polarity (MPQA, \citet{Wiebe2005AnnotatingEO}), semantic textual similarity (STS14, \citet{Agirre2014SemEval2014T1}), and SNLI \citep{Bowman2015ALA}. After unsupervised training, the encoder is fixed, and applied as a representation extractor on the 10 tasks. 

To compare the effect of different corpora, we also trained two models on Amazon Book Review dataset (without ratings) which is the largest subset of the Amazon Review dataset \citep{McAuley2015ImageBasedRO} with 142 million sentences, about twice as large as BookCorpus. 

Both training and evaluation of our models were conducted in PyTorch\footnote{http://pytorch.org/}, and we used SentEval\footnote{https://github.com/facebookresearch/SentEval} provided by \citet{Conneau2017SupervisedLO} to evaluate the transferability of our models. All the models were trained for the same number of iterations with the same batch size, and the performance was measured at the end of training for each of the models. 

\section{Related work and Comparison}
Table \ref{related-work} presents the results on 9 evaluation tasks of our proposed RNN-CNN models, and related work. The ``\textbf{small RNN-CNN}'' refers to the model with the dimension of representation as 1200, and the ``\textbf{large RNN-CNN}'' refers to that as 4800. The results of our ``large RNN-CNN'' model on SNLI is presented in Table \ref{snli}.
\begin{table}[h]
\fontsize{9}{12}\selectfont
\begin{center}
\begin{tabular}{l|c}
\hline
\hline
\textbf{Model} & \textbf{SNLI} (Acc \%) \\
\hline
\hline
\multicolumn{2}{c}{\emph{Unsupervised Transfer Learning}} \\
\hline
Skip-thoughts (\citeauthor{Vendrov2015OrderEmbeddingsOI}) & 81.5 \\
large RNN-CNN \emph{BookCorpus} & \textbf{81.7}\\
large RNN-CNN \emph{Amazon} & 81.5 \\
\hline
\hline
\multicolumn{2}{c}{\emph{Supervised Training}} \\
\hline
ESIM (\citeauthor{Chen2016EnhancingAC}) & 86.7 \\
DIIN (\citeauthor{Gong2017NaturalLI}) & \textbf{88.9} \\
\hline
\end{tabular}
\caption{We implemented the same classifier as mentioned in \citet{Vendrov2015OrderEmbeddingsOI} on top of the features computed by our model. Our proposed RNN-CNN model gets similar result on SNLI as skip-thoughts, but with much less training time.}
\label{snli}
\end{center}
\end{table}

Our work was inspired by analysing the \textbf{skip-thoughts} model \citep{Kiros2015SkipThoughtV}. The skip-thoughts model successfully applied this form of learning from the context information into unsupervised representation learning for sentences, and then, \citet{Ba2016LayerN} augmented the LSTM with proposed layer-normalisation (\textbf{Skip-thought+LN}), which improved the skip-thoughts model generally on downstream tasks. In contrast, \citet{Hill2016LearningDR} proposed the \textbf{FastSent} model which only learns source and target word embeddings and  is an adaptation of \textbf{Skip-gram} \citep{Mikolov2013DistributedRO} to sentence-level learning without word order information. 
\citet{Gan2017LearningGS} applied a CNN as the encoder, but still applied LSTMs for decoding the adjacent sentences, which is called \textbf{CNN-LSTM}.

Our \textbf{RNN-CNN} model falls in the same category as it is an encoder-decoder model. Instead of decoding the surrounding two sentences as in skip-thoughts, FastSent and the compositional CNN-LSTM, our model only decodes the subsequent sequence with a fixed length. Compared with the hierarchical CNN-LSTM, our model showed that, with a proper model design, the context information from the subsequent words is sufficient for learning sentence representations. Particularly, our proposed \textbf{small RNN-CNN} model runs roughly three times faster than our implemented skip-thoughts model\footnote{We reimplemented the skip-thoughts model in PyTorch, and it took roughly four days to train for only one epoch on BookCorpus.} on the same GPU machine during training.

Proposed by \citet{Radford2017LearningTG}, \textbf{BYTE m-LSTM} model uses a multiplicative LSTM unit \citep{Krause2016MultiplicativeLF} to learn a language model. This model is simple, providing next-byte prediction, but achieves good results likely due to the extremely large training corpus (Amazon Review data, \citet{McAuley2015ImageBasedRO}) that is also highly related to many of the sentiment analysis downstream tasks (domain matching).


We experimented with the Amazon
Book review dataset, the largest subset of the Amazon Review.
This subset is significantly smaller than the full Amazon Review dataset but twice as large as BookCorpus.
Our RNN-CNN model trained on the Amazon Book review dataset resulted in performance improvement on all single-sentence classification tasks relative to that achieved with training under BookCorpus. 

Unordered sentences are also useful for learning representations of sentences. \textbf{ParagraphVec} \citep{Le2014DistributedRO} learns a fixed-dimension vector for each sentence by predicting the words within the given sentence. However, after training, the representation for a new sentence is hard to derive, since it requires optimising the sentence representation towards an objective. \textbf{SDAE} \citep{Hill2016LearningDR} learns the sentence representations with a denoising auto-encoder model. 
Our proposed RNN-CNN model trains faster than SDAE does, 
and also because we utilised the sentence-level continuity as a supervision which SDAE doesn't, our model largely performs better than SDAE.

Another transfer approach is to learn a supervised discriminative classifier by distinguishing whether the sentence pair or triple comes from the same context. \citet{Li2014AMO} proposed a model that learns to classify whether the input sentence triplet contains three contiguous sentences. \textbf{DiscSent} \citep{Jernite2017DiscourseBasedOF} and \textbf{DisSent} \citep{Nie2017DisSentSR} both utilise the annotated explicit discourse relations, which is also good for learning sentence representations. It is a very promising research direction since the proposed models are generally computational efficient and have clear intuition, yet more investigations need to be done to augment the performance. 

Supervised training for transfer learning is also promising when a large amount of human-annotated data is accessible. \citet{Conneau2017SupervisedLO} proposed the \textbf{InferSent} model, which applies a bi-directional LSTM as the sentence encoder with multiple fully-connected layers to classify whether the hypothesis sentence entails the premise sentence in SNLI \citep{Bowman2015ALA}, and MultiNLI \citep{Williams2017ABC}. The trained model demonstrates a very impressive transferability on downstream tasks, including both supervised and unsupervised. 
Our RNN-CNN model trained on Amazon Book Review data in an unsupervised way has better results on supervised tasks than \textbf{InferSent} but slightly inferior results on semantic relatedness tasks. 
We argue that labelling a large amount of training data is time-consuming and costly, while unsupervised learning provides great performance at a fraction of the cost. It could potentially be leveraged to initialise or more generally augment the costly human labelling, and make the overall system less costly and more efficient. 

\section{Discussion}

In \citet{Hill2016LearningDR}, internal consistency is measured on five single sentence classification tasks (MR, CR, SUBJ, MPQA, TREC), MSRP and STS-14, and was found to be only above the ``acceptable'' threshold. They empirically showed that models that worked well on supervised evaluation tasks generally didn't perform well on unsupervised ones. This implies that we should consider supervised and unsupervised evaluations separately, since each group has higher internal consistency.

As presented in Table \ref{related-work}, the encoders that only sum over pretrained word vectors  perform better overall than those with RNNs on unsupervised evaluation tasks, including STS14. In recent proposed log-bilinear models, such as \textbf{FastSent} \citep{Hill2016LearningDR} and \textbf{SiameseBOW} \citep{Kenter2016SiameseCO}, the sentence representation is composed by summing over all word representations, and the only tunable parameters in the models are word vectors. These resulting models perform very well on unsupervised tasks. By augmenting the pretrained word vectors with a weighted averaging process, and removing the top few principal components, which mainly encode frequently-used words, as proposed in \citet{Arora2017ASB} and \citet{Mu2017AllbuttheTopSA}, the performance on the unsupervised evaluation tasks gets even better. Prior work suggests that incorporating word-level information helps the model to perform better on cosine distance based semantic textual similarity tasks. 

Our model predicts all words in the target sequence at once, without an autoregressive process, and ties the word embedding layer in the encoder with the prediction layer in the decoder, which explicitly uses the word vectors in the target sequence as the supervision in training. Thus, our model incorporates the word-level information by using word vectors as the targets, and it improves the model performance on STS14 compared to other RNN-based encoders.

\citet{Arora2017ASB} conducted an experiment to show that the word order information is crucial in getting better results on supervised tasks. In our model, the encoder is still an RNN, which explicitly utilises the word order information. We believe that the combination of encoding a sentence with its word order information and decoding all words in a sentence independently inherently leverages the benefits from both log-linear models and RNN-based models.


\section{Conclusion} 

Inspired by learning to exploit the contextual information present in adjacent sentences, we proposed an asymmetric encoder-decoder model with a suite of techniques for improving context-based unsupervised sentence representation learning. Since we believe that a simple model will be faster in training and easier to analyse, we opt to use simple techniques in our proposed model, including 1) an RNN as the encoder, and a predict-all-words CNN as the decoder, 2) learning by inferring subsequent contiguous words, 3) mean+max pooling, and 4) tying word vectors with word prediction. With thorough discussion and extensive evaluation, we justify our decision making for each component in our RNN-CNN model. In terms of the performance and the efficiency of training, we justify that our model is a fast and simple algorithm for learning generic sentence representations from unlabelled corpora. Further research will focus on how to maximise the utility of the context information, and how to design simple architectures to best make use of it.


\clearpage
\bibliography{acl2018}
\bibliographystyle{acl_natbib}

\clearpage
\appendix

\section{Supplemental Material}
\label{sec:supplemental}
Table \ref{architecture} presents the effect of hyperparameters.

\begin{table*}[t]
\fontsize{9}{12}\selectfont
\begin{center}
\begin{tabular}{K{0.4cm} K{0.8cm} | K{0.8cm} K{2.8cm}| K{0.5cm} | K{0.8cm} K{1.0cm} K{1.1cm} | K{1.2cm} | K{0.4cm} K{0.8cm}  }
\hline
\multicolumn{2}{c|}{\textbf{Encoder}} & \multicolumn{2}{c|}{\textbf{Decoder}} & \multirow{2}{*}{\textbf{Hrs}} & \multirow{2}{*}{\textbf{SICK-R}} &  \multirow{2}{*}{\textbf{SICK-E}} & \multirow{2}{*}{\textbf{STS14}} & \textbf{MSRP} & \multirow{2}{*}{\textbf{SST}} & \multirow{2}{*}{\textbf{TREC}}   \\ \cline{1-4} 
type & dim & type & dim & & & & & (Acc/F1) & & \\
\hline
\hline
\multicolumn{11}{c}{\textbf{Dimension of Sentence Representation: 1200}} \\
 \hline
 \hline
 \multirow{4}{*}{RNN} & \multirow{4}{*}{2x300} & CNN & 600-1200-300 & 20 & 0.8530 & 82.6 & 0.58/0.56 & \underline{\textbf{75.6}}/\textbf{82.9} & 82.8 & \textbf{89.2} \\
  &  & CNN$^\dag$ & 600-1200-300 & 21 & 0.8515 & 82.7 & 0.58/0.56 & 75.3/82.5 & \textbf{82.9} & 85.2 \\
  &  & CNN(10) & 600-1200-300 & 11 & 0.8474 & 82.9 & 0.57/0.55 & 74.2/81.6 & 82.8 & 88.0 \\
  &  & CNN(50) & 600-1200-300 & 27 & 0.8533 & 82.5 & 0.57/0.55 & 74.7/82.2 & 81.5 & 86.2 \\
 \hline
 RNN & 2x300 & RNN & 600 & 26 & 0.8530 & 82.6 & 0.51/0.50 & 74.1/81.7 & 81.0 & 89.0 \\
 CNN & 4x300$^\S$ & CNN & 600-1200-300 & 8 & 0.8117 & 80.5 & 0.44/0.42 & 72.7/80.7 & 78.4 & 85.0 \\
 \hline
 \multirow{2}{*}{RNN} & \multirow{2}{*}{2x300} & CNN & 600-1200-2400-300 & 28 & \textbf{0.8570} & \textbf{84.0} & 0.58/0.56 &  74.3/81.5 & 82.8 & 88.2 \\ 
  & & CNN & 1200-2400-300 & 27 & 0.8541 & 83.0 & \underline{\textbf{0.59}}/\underline{\textbf{0.57}} & 74.3/82.2 & \textbf{82.9} & 89.0 \\
 \hline
 \hline
 \multicolumn{11}{c}{\textbf{Dimension of Sentence Representation: 2400}} \\
 \hline
 \hline
 RNN & 2x600 & CNN & 600-1200-300 & 25 & 0.8631 & 83.9 & \textbf{0.58}/\textbf{0.55} & \textbf{74.7}/\textbf{83.1} & 83.4 & \textbf{90.2} \\ 
 \hline
 RNN & 2x600 & RNN & 600 & 32 & \textbf{0.8647} & \textbf{84.2} & 0.52/0.51 & 74.0/81.2 & \underline{\textbf{84.2}} & 87.6 \\
 CNN & 3x800$^\ddag$ & RNN & 600 & 8 & 0.8132 & - & - & 71.9/81.9 & - & 86.6 \\
 \hline
 \hline
 \multicolumn{11}{c}{\textbf{Dimension of Sentence Representation: 4800}} \\
 \hline
 \hline
 RNN & 2x1200 & CNN & 600-1200-300 & 34 & \underline{\textbf{0.8698}} & \underline{\textbf{85.2}} & \underline{\textbf{0.59}}/\underline{\textbf{0.57}} & \textbf{75.1}/\underline{\textbf{83.2}} & \textbf{84.1} & \underline{\textbf{92.2}} \\ 
 \hline
 \multicolumn{4}{c|}{Skip-thought \citep{Kiros2015SkipThoughtV}} & 336 & 0.8584 & 82.3 & 0.29/0.35 & 73.0/82.0 & 82.0 & \underline{\textbf{92.2}} \\
 \multicolumn{4}{c|}{Skip-thought+LN \citep{Ba2016LayerN}} & 720 & 0.8580 & 79.5 & 0.44/0.45 & - & 82.9 & 88.4 \\
\hline
\hline
\end{tabular}
\end{center}
\caption{\textbf{Architecture Comparison}. As shown in the table, our designed asymmetric RNN-CNN model (row 1,9, and 12) works better than other asymmetric models (CNN-LSTM, row 11), and models with symmetric structure (RNN-RNN, row 5 and 10). In addition, with larger encoder size, our model demonstrates stronger transferability. The default setting for our CNN decoder is that it learns to reconstruct 30 words right next to every input sentence. ``CNN(10)'' represents a CNN decoder with the length of outputs as 10, and ``CNN(50)'' represents it with the length of outputs as 50. ``\dag'' indicates that the CNN decoder learns to reconstruct next sentence. ``\ddag'' indicates the results reported in \citeauthor{Gan2017LearningGS} as \emph{future predictor}. The CNN encoder in our experiment, noted as ``\S'', was based on AdaSent in \citeauthor{Zhao2015SelfAdaptiveHS} and \citeauthor{Conneau2017SupervisedLO}. \textbf{Bold} numbers are best results among models at same dimension, and underlined numbers are best results among all models. For STS14, the performance measures are Pearson's and Spearman's score. For MSRP, the performance measures are accuracy and F1 score.}
\label{architecture}
\end{table*}

\subsection{Decoding Sentences vs. Decoding Sequences}
Given that the encoder takes a sentence as input, decoding the next sentence versus decoding the next fixed length window of contiguous words is conceptually different. This is because decoding the subsequent fixed-length sequence might not reach or might go beyond the boundary of the next sentence. Since the CNN decoder in our model takes a fixed-length sequence as the target, when it comes to decoding sentences, we would need to zero-pad or chop the sentences into a fixed length. As the transferability of the models trained in both cases perform similarly on the evaluation tasks (see rows 1 and 2 in Table \ref{architecture}), we focus on the simpler predict-all-words CNN decoder that learns to reconstruct the next window of contiguous words.

\subsection{Length of the Target Sequence $T$}
We varied the length of target sequences in three cases, which are 10, 30 and 50, and measured the performance of three models on all tasks. As stated in rows 1, 3, and 4 in Table \ref{architecture}, decoding short target sequences results in a slightly lower Pearson score on SICK, and decoding longer target sequences lead to a longer training time. In our understanding, decoding longer target sequences leads to a harder optimisation task, and decoding shorter ones leads to a problem that not enough context information is included for every input sentence. A proper length of target sequences is able to balance these two issues. The following experiments set subsequent 30 contiguous words as the target sequence.

\subsection{RNN Encoder vs. CNN Encoder}
The CNN encoder we built followed the idea of AdaSent \citep{Zhao2015SelfAdaptiveHS}, and we adopted the architecture proposed in \citep{Conneau2017SupervisedLO}. The CNN encoder has four layers of convolution, each followed by a non-linear activation function. At every layer, a vector is calculated by a global max-pooling function over time, and four vectors from four layers are concatenated to serve as the sentence representation. We tweaked the CNN encoder, including different kernel size and activation function, and we report the best results of CNN-CNN model at row 6 in Table \ref{architecture}. 

Even searching over many hyperparameters and selecting the best performance on the evaluation tasks (overfitting), the CNN-CNN model performs poorly on the evaluation tasks, although the model trains much faster than any other models with RNNs (which were not similarly searched). The RNN and CNN are both non-linear systems, and they both are capable of learning complex composition functions on words in a sentence. We hypothesised that the explicit usage of the word order information will augment the transferability of the encoder, and constrain the search space of the parameters in the encoder. The results support our hypothesis.

The \emph{future predictor} in \cite{Gan2017LearningGS} also applies a CNN as the encoder, but the decoder is still an RNN, listed at row 11 in Table \ref{architecture}. Compared to our designed CNN-CNN model, their CNN-LSTM model contains more parameters than our model does, but they have similar performance on the evaluation tasks, which is also worse than our RNN-CNN model. 

\subsection{Dimensionality}

Clearly, we can tell from the comparison between rows 1, 9 and 12 in Table \ref{architecture}, increasing the dimensionality of the RNN encoder leads to better transferability of the model.

Compared with RNN-RNN model, even with double-sized encoder, the model with CNN decoder still runs faster than that with RNN decoder, and it slightly outperforms the model with RNN decoder on the evaluation tasks.

At the same dimensionality of representation with Skip-thought and Skip-thought+LN, our proposed RNN-CNN model performs better on all tasks but TREC, on which our model gets similar results as other models do.

Compared with the model with larger-size CNN decoder, apparently, we can see that larger encoder size helps more than larger decoder size does (rows 7,8, and 9 in Table \ref{architecture}). 

In other words, an encoder with larger size will result in a representation with higher dimensionality, and generally, it will augment the expressiveness of the vector representation, and the transferability of the model.

\section{Experimental Details}
Our \textbf{small} RNN-CNN model has a bi-directional GRU as the encoder, with 300 dimension each direction, and the \textbf{large} one has 1200 dimension GRU in each direction. The batch size we used for training our model is 512, and the sequence length for both encoding and decoding are 30. The initial learning rate is $0.0005$, and the Adam optimiser \citep{Kingma2014AdamAM} is applied to tune the parameters in our model.

\section{Results including supervised task-dependent models}
Table \ref{allrelated} contains all supervised task-dependent models for comparison.

\begin{table*}[!t]
\fontsize{9}{12}\selectfont
\begin{center}
\begin{tabular} {l| K{0.4cm} | K{1.2cm} K{1.1cm} K{1.0cm} K{1.1cm} | K{0.6cm} K{0.5cm} K{0.5cm} K{0.5cm} K{0.6cm} K{0.6cm}}
\Xhline{2pt}
\textbf{Model} & \textbf{Hrs} & \textbf{SICK-R} & \textbf{SICK-E} & \textbf{STS14} & \textbf{MSRP} & \textbf{TREC} & \textbf{MR} & \textbf{CR} & \textbf{SUBJ} & \textbf{MPQA} & \textbf{SST} \\
\Xhline{2pt}
Measurement  & & $r$ & Acc. & $r$/$\rho$ & Acc./F1 & \multicolumn{6}{c}{Accuracy}\\
\hline
\multicolumn{12}{r}{\emph{Unsupervised training with unordered sentences}} \\
\hline
ParagraphVec & 4 & - & - & 0.42/0.43 & 72.9/81.1 & 59.4 & 60.2 & 66.9 & 76.3 & 70.7 & - \\
word2vec BOW & 2 & 0.8030 & 78.7 & 0.65/\textbf{0.64} & 72.5/81.4 & 83.6 & 77.7 & \textbf{79.8} & 90.9 & \textbf{88.3} &  79.7 \\
fastText BOW & - & 0.8000 & 77.9 & 0.63/0.62 & 72.4/81.2 & 81.8 & 76.5 & 78.9 & \textbf{91.6} & 87.4 & 78.8 \\
SIF (GloVe+WR) & - & \textbf{0.8603} & \textbf{84.6} & \textbf{0.69}/  -  & - / - & - & - & - & - & - & \textbf{82.2} \\
GloVe BOW & - & 0.8000 & 78.6 & 0.54/0.56 & 72.1/80.9 & 83.6 & \textbf{78.7} & 78.5 & \textbf{91.6} & 87.6 & 79.8 \\
SDAE & 72 & - & - & 0.37/0.38 & \textbf{73.7}/80.7 & 78.4 & 74.6 & 78.0 & 90.8 & 86.9 & -  \\
\Xhline{2pt}

\hline
\multicolumn{12}{r}{\emph{Unsupervised training with ordered sentences - BookCorpus}} \\
\hline
FastSent & 2 & - & - & \textbf{0.63}/\textbf{0.64} & 72.2/80.3 & 76.8 & 70.8 & 78.4 & 88.7 & 80.6 & - \\
FastSent+AE & 2 & - & - & 0.62/0.62 & 71.2/79.1 & 80.4 & 71.8 & 76.5 & 88.8 & 81.5 & - \\
\hline
Skip-thought & 336 & 0.8580 & 82.3 & 0.29/0.35 & 73.0/82.0& 92.2 & 76.5 & 80.1 & 93.6 & 87.1 & 82.0 \\
Skip-thought+LN & 720 & 0.8580 & 79.5 & 0.44/0.45 & - & 88.4 & 79.4 & \textbf{83.1} & 93.7 & 89.3 & 82.9 \\
combine CNN-LSTM & - & 0.8618 & - & - & \underline{\textbf{76.5}}/\underline{\textbf{83.8}} & \underline{\textbf{92.6}} & 77.8 & 82.1 & 93.6 & \textbf{89.4} & - \\
\hline
\textbf{\emph{small RNN-CNN}}\dag & 20 & 0.8530 & 82.6  & 0.58/0.56 & 75.6/82.9 & 89.2 & 77.6 & 80.3 & 92.3 & 87.8 & 82.8 \\
\textbf{\emph{large RNN-CNN}}\dag & 34 & \textbf{0.8698} & \textbf{85.2} & 0.59/0.57 & 75.1/83.2 & 92.2 & \textbf{79.7} & 81.9 & \textbf{94.0} & 88.7 & \textbf{84.1} \\
\Xhline{2pt}
\hline
\multicolumn{12}{r}{\emph{Unsupervised training with ordered sentences - Amazon Book Review}} \\
\hline
\textbf{\emph{small RNN-CNN}}\dag & 21 & 0.8476 & 82.7 & \textbf{0.53}/\textbf{0.53} & 73.8/81.5 & 84.8 & 83.3 & 83.0 & 94.7 & 88.2 & 87.8 \\
\textbf{\emph{large RNN-CNN}}\dag & 33 & \textbf{0.8616} & \textbf{84.3} & 0.51/0.51 & \textbf{75.7}/\textbf{82.8} & \textbf{90.8} & 85.3 & 86.8 & \underline{\textbf{95.3}} & \textbf{89.0} & \underline{\textbf{88.3}} \\
\hline
\hline
\multicolumn{12}{r}{\emph{Unsupervised training with ordered sentences - Amazon Review}} \\
\hline
BYTE m-LSTM  & 720 & 0.7920 & - & - & 75.0/\textbf{82.8} & - & \underline{\textbf{86.9}} & \underline{\textbf{91.4}} & 94.6 & 88.5 & - \\
\Xhline{2pt}
\hline
\multicolumn{12}{r}{\emph{Supervised training - Transfer learning}} \\
\hline
DiscSent & 8 & - & - & - & 75.0/ -  & 87.2 & - & - & 93.0 & - & - \\
DisSent Books 8 & - & 0.8170 & 81.5 & - & -/ -  & 87.2 & \textbf{82.9} & 81.4 & \textbf{93.2} & 90.0 & 80.2 \\
\hline
CaptionRep BOW & 24 & - & - & 0.46/0.42 & -  & 72.2 & 61.9 & 69.3 & 77.4 & 70.8 & - \\
DictRep BOW & 24 & - & - & 0.67/\underline{\textbf{0.70}} & 68.4/76.8 & 81.0 & 76.7 & 78.7 & 90.7 & 87.2 & - \\
InferSent(SNLI) & $<$24 & \underline{\textbf{0.8850}} & 84.6 & 0.68/0.65 & 75.1/82.3 & \textbf{88.7} & 79.9 & 84.6 & 92.1 & 89.8 & 83.3\\
InferSent(AllNLI)  & $<$24 & 0.8840 & \underline{\textbf{86.3}} & \underline{\textbf{0.70}}/0.67 & \textbf{76.2}/\textbf{83.1} & 88.2 & 81.1 & \textbf{86.3} & 92.4 & \underline{\textbf{90.2}} & \textbf{84.6} \\
\Xhline{2pt}
\hline
\end{tabular}
\end{center}
\caption{\textbf{Related Work and Comparison.} As presented, our designed asymmetric RNN-CNN model has strong transferability, and is overall better than existing unsupervised models in terms of fast training speed and good performance on evaluation tasks. ``\dag''s refer to our models, and ``\textbf{small}/\textbf{large}'' refers to the dimension of representation as 1200/4800. \textbf{Bold} numbers are the best ones among the models with same training and transferring setting, and \underline{underlined} numbers are best results among all transfer learning models. The training time of each model was collected from the paper that proposed it.}
\label{allrelated}
\end{table*}

\end{document}